\documentclass[twoside]{memoir}
\usepackage{nus-el-ht}
\usepackage{cite}
\usepackage{hyperref}
\hypersetup{
    colorlinks=true,
    linkcolor=blue,
    filecolor=magenta,   
    urlcolor=cyan,
    citecolor=black,
    pdftitle={Overleaf Example},
    pdfpagemode=FullScreen,
    }

\usepackage{float}

\usepackage{lmodern}
\PassOptionsToPackage{hyphens}{url}\usepackage{hyperref}
\DisemulatePackage{setspace}\usepackage{setspace}
\expandafter\def\expandafter\quote\expandafter{\quote\singlespacing}
\usepackage{tipa}

\usepackage{qtree}\usepackage{tikz} 
\usepackage{tikz-qtree}
	\tikzset{every tree node/.style={align=center,anchor=north}}
\usepackage{expex} 
	\lingset{Everyex=\singlespace}

\nouppercaseheads
\title{Neural Style Transfer for Synthesising a Dataset of Ancient Egyptian Hieroglyphs
}
\author{Lewis Matheson Creed}
\supervisor {supervised by Dr Vithya Yogarajan, Professor Gillian Dobbie \\and Dr Jennifer Hellum}
\degree{Bachelor of Advanced Science with Honours / Bachelor of Arts in Computer Science, History and Ancient History \& Classical Studies}
\faculty{Faculty of Science}
\dept{Computer Science}
\date{11 November 2024}

\begin{document}
\frontmatter


\maketitle

\chapter{Acknowledgement}
First, I would like to thank my supervisors, Dr Vithya Yogarajan, Professor Gillian Dobbie and Dr Jennifer Hellum, for their guidance, support and feedback over the last 12 months. I would also like to thank Di Zhao, PhD candidate at the University of Auckland, for shaping the direction of this project by suggesting I explore Neural Style Transfer. I'd like to thank the Machine Learning Group at The University of Auckland for their time and feedback at my Honours presentation. Last, thank you to Derek Lim and Peter Tan for their Honours thesis template\cite{template}.

\begin{abstract}
  The limited availability of training data for low-resource languages makes applying machine learning (ML) techniques challenging. Ancient Egyptian is one such language with few resources. For example, the only publicly available dataset for ML classification of ancient Egyptian hieroglyphs represents less than 22.5\% of the logographic language’s characters. However, innovative applications of data augmentation methods, such as Neural Style Transfer (NST), could overcome these barriers. This paper presents a novel method for generating datasets of ancient Egyptian hieroglyphs by applying NST to a digital typeface. Experimental results found that image classification models trained on NST-generated examples and photographs demonstrate equal performance and transferability to real unseen images of hieroglyphs.
\end{abstract}

\tableofcontents


\mainmatter

\chapter{Introduction}

Ancient Egyptian is one of the oldest written languages in human history, dating back over 5000 years \cite{britannica2024}. Despite falling out of use inside Egypt thousands of years ago and only being decipherable again in the late 18th century, the ancient Egyptian language maintains a widespread, international popular interest due to its beautiful logographic hieroglyphs \cite{britishmuseum2024rosetta}. However, very few specialist Egyptologists and hobby enthusiasts can translate ancient Egyptian writing, and doing so often requires the assistance of dictionaries and other resources because of the complexities of the language. Thus, developing Machine Learning (ML) translation tools for ancient Egyptian could help make the language more accessible, opening up new and interactive learning opportunities for millions of tourists in Egypt \cite{xinhua2024egypt} and museum-goers worldwide, not to mention students of Egyptology and other academics working with the language. However, general progress towards ML-powered translations of ancient Egyptian has been slowed primarily by two obstacles.

First, the logographic nature of the ancient Egyptian language complicates the ML translation task. When reading ancient Egyptian texts, the hieroglyphic characters must first be transliterated from symbols into phonetic values before being translated. Human translators are usually assisted in the classification process by Gardiner’s sign list \cite{gardiner1957hieroglyphic}, a reference resource for the transliterated values of common Egyptian hieroglyphic characters. Gardiner’s sign list \cite{gardiner1957hieroglyphic} encodes each unique hieroglyphic character into 765 individual classes, one for each unique sign, and these are grouped into 26 broader superclasses. For example, the “owl” hieroglyph has the transliterated value of “m” and is classified as a G17 class of hieroglyph\cite{gardiner1957hieroglyphic}. G refers to the superclass for Birds and 17 is the number that character is given within the superclass. ML research uses Gardiner’s labelling conventions to annotate and classify ancient Egyptian hieroglyphs \cite[\dots]{franken2013automatic, aneesh2024exploring}.

\begin{figure}[H]
    \centering
    \includegraphics[width=0.8\textwidth]{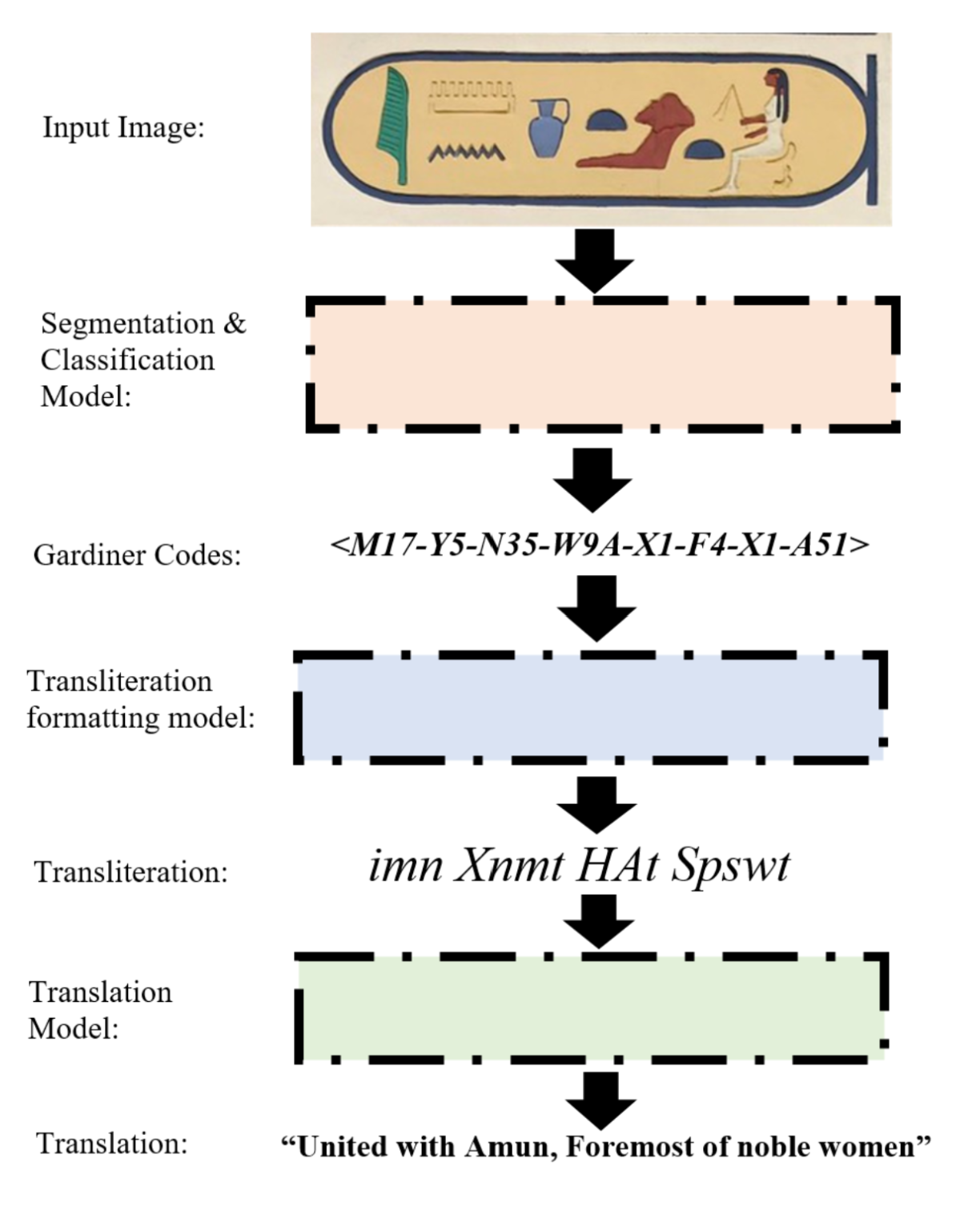}\label{fig:hatshepsut_translation}
    \caption{Diagram of a hypothetical automatic ancient Egyptian translation pipeline. The ancient Egyptian writing depicts the Son of Re name, or "official name," for Queen Hatshepsut of the 18th Dynasty during the New Kingdom period. Photo sourced from \cite{guestnz2024hamilton}.}
    
\end{figure}

Due to this transliteration step, ancient Egyptian texts cannot be input directly as text into existing ML translation tools. Therefore, the first step of a hypothetical translation pipeline involves ML image classification to identify hieroglyphs present in the text so that they can be transliterated (see Figure \ref{fig:hatshepsut_translation}).

There are currently no ML translation tools, such as Google Translate \cite{google2024translate}, available for ancient Egyptian. There are a handful of apps such as Google's Fabricius \cite{google2024fabricius} which can identify individual hieroglyphs, or characters, but these are not open source. Therefore, focusing on this ML classification of individual hieroglyphs is where the current research frontier stands \cite[\dots]{barucci2021deep,aneesh2024exploring}. Although a handful of papers have begun to explore classifying multiple hieroglyphs in an image via segmentation \cite{lion2024unsupervised,lombardi2024localisation,cucci2024hyperspectral}.

The second obstacle is that ancient Egyptian is a low-resource language, meaning there is an overall limited availability of training data for ML models. Although a substantial amount of ancient Egyptian text has survived, there is little need from within Egyptology to develop ML translation software for the language, as new texts are rarely discovered. As such, the effort required to manually extract, annotate, and categorise data from these extant texts is not justifiable. Reflecting this deficit, there is currently only one open source dataset for ML classification of ancient Egyptian hieroglyphs: the Unas dataset published by Franken and Gemert in 2013 \cite{franken2013automatic}. It contains 4,210 manually annotated images taken from photographs of the Unas Pyramid texts in Saqqara \cite{franken2013automatic}. However, this dataset only includes 171 classes of hieroglyphs, representing 22.5\% of the total number of common ancient Egyptian characters \cite{franken2013automatic,gardiner1957hieroglyphic}. Thus, while the Unas dataset \cite{franken2013automatic} is instrumental, a more complete dataset is needed to classify more of the language.

Recent innovations in ML have introduced the possibility of automated image generation, such as Neural Style Transfer (NST) \cite{gatys2015neural}. NST is a ML-based art generation method first published in 2015 by Gatys et al., which takes two input images and synthesises an output image where the \textit{style} of one image (i.e. colours and textures) is applied to the \textit{content} of the other image, retaining its underlying form \cite{gatys2015neural}. Since 2015, research has begun to explore utilising NST as a data augmentation method for injecting heterogeneous noise, or variety, into homogeneous data with promising results \cite[\dots]{mikolajczyk2018augmentation,mikolajczyk2019style,darma2020balinese,xiao2021progressive}. Training a model on heterogeneous data helps reduce overfitting, where a classification model becomes too familiar with the data it is trained on; resulting in a reduced ability to classify unseen data (transferability). Since ML research aims to train intelligent and generalisable computer systems, overfitting is to be avoided. Data augmentation is also helpful for creating additional training images for low resource datasets with small class sizes \cite{darma2020balinese}. 

Although a publicly available and complete dataset (i.e. containing all 765 classes) for ancient Egyptian hieroglyphs does not exist, several open-source digital typefaces exist, such as J-Sesh \cite{jsesh2024jsesh} and the Unicode standard \cite{unicode2024hieroglyphs}. These typefaces are complete because they include at least one example for every common hieroglyphic character \cite{gardiner1957hieroglyphic}. As such, they could be viewed as low-resource datasets with very small class sizes. Since real ancient Egyptian hieroglyphic characters were crafted in numerous art styles, transferring those styles from photographs onto such sterile fonts is an intuitive usage of NST augmentation. Furthermore, since such an approach could be automated to synthesise unlimited numbers of example images, it has potential to entirely overcome the problem of ancient Egyptian being a low-resource language.

\section{Research Questions}

Therefore, this research seeks to investigate this novel methodology of applying NST to a digital typeface, generating a dataset. The generated NST dataset will then be evaluated to see if it is comparably effective to the existing Unas dataset. The research questions are as follows:

\textbf{RQ1:} \textit{Can NST be used as a data augmentation method on a digital typeface of ancient Egyptian hieroglyphs for ML classification?}

\textbf{RQ2:} \textit{Is such an NST augmented digital typeface comparable to an existing photographic dataset for ML classification?}

To address these research questions, I formulated a method for applying NST as a data augmentation method to a typeface. This method was then used to synthesise a novel NST dataset of ancient Egyptian hieroglyphs. Then the quality of this NST dataset was evaluated by comparing the transferability and performance of a ML classification model trained on the dataset against two other datasets with other types of images (photographs and non-NST-augmented typeface characters).

\chapter{Related Work}

\section{Neural Style Transfer: A Data Augmentation Tool
}

Since its conception, NST has been deployed as a data augmentation tool in many different visual contexts to introduce heterogeneity and synthesise data \cite{atarsaikhan2017neural ,mikolajczyk2018augmentation,mikolajczyk2019style,darma2020balinese,xiao2021progressive,li2023cross ,mumuni2024survey}. NST is a versatile tool that can produce stylised and photorealistic outputs depending on the parameters and inputs \cite{mumuni2024survey}.

NST has been applied to typefaces, but only to synthesise other stylised typefaces, not to create a dataset or generate a typeface made of photorealistic materials. In 2017, Atarsaikhan et al. \cite{atarsaikhan2017neural} explored merging different patterns, languages and pre-existing fonts via NST to generate new typefaces for Latin, Arabic, Japanese, Korean and Cyrillic scripts. More recent research in this direction, such as that by Li et al. (2022), has seen purpose-built neural networks made for transferring the style of typefaces across different scripts \cite{li2023cross}. The languages considered by existing research in this area are not low-resource, although Li et al. experimented with a logographic script (Chinese), a language with functionally similar characters to ancient Egyptian hieroglyphics \cite{li2023cross}.

Regarding low-resource problem domains where photorealistic outputs are required, NST has been used to synthesise datasets. For example, in 2020, NST was used by Darma et al. \cite{darma2020balinese} to expand a very small dataset of 45 images of Balinese carvings into a larger dataset of  4095 augmented examples. They then trained eight models using a MobileNet CNN (Convolutional Neural Network) architecture, each with a different training dataset variation \cite{darma2020balinese}. They then tested these models on a dataset of 191 unseen photos of Balinese carvings \cite{darma2020balinese}. Their results found that models trained on the NST synthesised datasets scored 12.5\% higher than models trained on un-augmented data \cite{darma2020balinese}. Furthermore, they also found that NST data augmentation outperformed affine data augmentation by 4.7\% and that using NST and affine data augmentation boosted the accuracy score by an additional 3.7\% \cite{darma2020balinese}. Other research in low-resource domains concurs with these findings, suggesting NST can be used to successfully synthesise datasets \cite{mikolajczyk2019style,xiao2021progressive}.

\section{Machine Learning Classification of Ancient Egyptian Hieroglyphs}

In the years following Franken and Gemert’s (2013) \cite{franken2013automatic} pioneering paper, numerous papers \cite[\dots]{duque2017deciphering, elnabawy2018image,
barucci2021deep,
moustafa2022scriba,
mohsen2023aegyptos,
sobhy2023translator,
guidi2023segmentation,
lion2024unsupervised,
aneesh2024exploring,
cucci2024hyperspectral,
lombardi2024localisation} have gone on to prove ML models, primarily CNNs, are capable of accurately classifying ancient Egyptian hieroglyphs. These papers have also relied in-part or heavily on the Unas dataset \cite{duque2017deciphering, elnabawy2018image,
barucci2021deep,
moustafa2022scriba,
mohsen2023aegyptos,
sobhy2023translator,
guidi2023segmentation,
lion2024unsupervised}. There is also a open source photographic dataset used for testing ML segmentation of hieroglyphs \cite{custer2024} but this dataset contains scenes with many characters and are unannotated, so it is unsuitable for ML classification training.

One of the most cited papers to explore this research area is by Barucci et al. (2021) \cite{barucci2021deep}. The authors created a purpose-built Convolutional Neural Network (CNN) to classify ancient Egyptian hieroglyphs called GlyphNet \cite{barucci2021deep}. The architecture of GlyphNet is based on the Xception \cite{barucci2021deep} CNN but was tailored to address the problem domain of ancient Egyptian hieroglyphs. To evaluate the effectiveness of GlyphNet, it was trained alongside three other CNNs that were mostly trained the same subset of the Unas dataset containing 40 classes of hieroglyphs\cite{franken2013automatic}. This subset was also supplemented by some additional example images collected by the authors from other sources \cite{barucci2021deep}. Their experiments found that GlyphNet outperformed the other CNNs in terms of accuracy, recall, precision, and F1 score, with very high results, scoring above 96\% \cite{barucci2021deep}.

Since 2021, research has begun to look towards compiling new datasets. First, Moustafa et al. (2022) \cite{moustafa2022scriba} published a paper outlining an ML translation app they were working on called \textit{Scriba}, which was trained on a data set that was a combination of the Unas Pyramid Text dataset \cite{franken2013automatic} and 6,000 additional images taken from archaeological sites, cartouche lists and Google images \cite{moustafa2022scriba}. In total, their dataset contained 87 classes of hieroglyphs. In 2023, the same research group followed up by outlining a new ML translation app for ancient Egyptian called \textit{Aegyptos}, relying on a new dataset containing 60,000 images and 1,072 classes of hieroglyphs, which included 994 images from the Unas dataset \cite{mohsen2023aegyptos} and their 6,000 images originally compiled for \textit{Scriba}\cite{moustafa2022scriba}.

In 2024, Aneesh et al. \cite{aneesh2024exploring} published a paper outlining a novel dataset they developed containing 4,578 images across 763 classes of hieroglyphs. Their method involved hand-drawing each hieroglyph in Gardiner’s sign list five times \cite{aneesh2024exploring}. After this, they applied affine, noise, and resolution data augmentations, expanding their dataset to 59,514 images for training purposes. They trained and tested five CNNs on their dataset, with F1 scores ranging between 78\% and 98\%.

Since the datasets used in  \cite{moustafa2022scriba,mohsen2023aegyptos,aneesh2024exploring} have yet to be made open source, there remains a research gap for a complete open-source dataset for ancient Egyptian hieroglyphics. Furthermore, there needs to be an established and shared benchmark for ancient Egyptian hieroglyphs. This is because each research paper  \cite{duque2017deciphering, elnabawy2018image,
barucci2021deep,
moustafa2022scriba,
mohsen2023aegyptos,
sobhy2023translator,
guidi2023segmentation,
lion2024unsupervised} uses a different mixture of datasets and quantity of classes, making it difficult to compare approaches and replicate results.

Furthermore, much of the research conducted thus far demonstrates high performance when tested on the same type of data they were trained on \cite{barucci2021deep}. However, there is a need to assess whether these models are transferable to classify real unseen hieroglyphs. For example, the Unas dataset \cite{franken2013automatic}, which much of the research depends on, is highly homogeneous in appearance, where the characters are written in one style and sourced from one location and period. Figure \ref{fig:variety_styles} shows some randomly selected images of the same character (G17) taken from different archaeological sites in Egypt, highlighting the heterogeneity of hieroglyphic art styles. In addition, ancient Egyptian scribes and artisans did not draw hieroglyphic characters with a pen/pencil like modern people do, like in \cite{gardiner1957hieroglyphic} and \cite{aneesh2024exploring}. Instead, they used a separate cursive script called hieratic \cite{hieratic}. No work has yet explored classifying hieratic scripts, and no datasets exist for that style of ancient Egyptian text.

To help address these problems, the datasets and code used in my research will be released publicly, with the hope that they will assist future researchers. Likewise, an additional transferability test will be conducted with unseen images of real hieroglyphs.

\chapter{Methodology}

This section outlines the various datasets considered, how they were created, and the experimental conditions set up to answer the research questions.

\section{Datasets Overview}

Four datasets were compiled to answer the research questions, three of which were used to train classification models. The first dataset, called the Unas* dataset, used a subset of the photographic images from the Unas dataset \cite{franken2013automatic}. The second dataset was synthesised via NST, combining characters from the J-Sesh typeface and photos of real hieroglyphs (the G17 dataset). The third dataset consisted of duplicated characters from the J-Sesh typeface \cite{jsesh2024jsesh}. These three datasets were set up with identical class distributions of 3,107 images across 34 classes of hieroglyphs, with a median class size of 71 (Figure \ref{fig:class_distribution}). A glossary of the various datasets used is in the appendix of this report \ref{tab:6}.

\begin{figure}[H]
    \centering
    \includegraphics[width=0.9\textwidth]{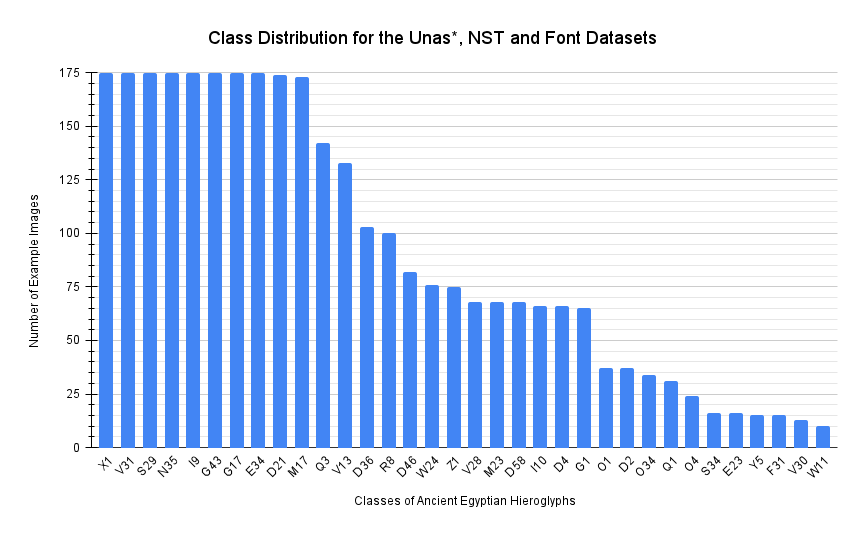}
    \caption{Class distribution of example images inside the Unas* dataset after the specified adjustments were made to the original Unas dataset. This same distribution was used for this paper's NST and Font datasets. The median class size is 71.}
    \label{fig:class_distribution}
\end{figure}

The dataset distribution in Figure \ref{fig:class_distribution} was determined by first taking the Unas dataset \cite{franken2013automatic} as a starting point and removing any classes with less than 10 example images to allow for a 10\% coefficient variation. Then, any classes not considered by Barucci et al. \cite{barucci2021deep} were removed so their model could be used as an independent reference point for the experiments presented here. These restrictions reduced the number of classes in the dataset from 171 to 34. To further balance the dataset, I decided to cap the maximum examples for a class to 175 to help reduce class imbalance. Any excess images were removed with a Python script using the in-built random sample function. The datasets used are summarised in Figure \ref{fig:dataset_breakdown}. There is also a glossary in the appendix that summarises the various datasets used \ref{tab:6}.

\begin{figure}[H]
    \centering
    \includegraphics[width=0.9\textwidth]{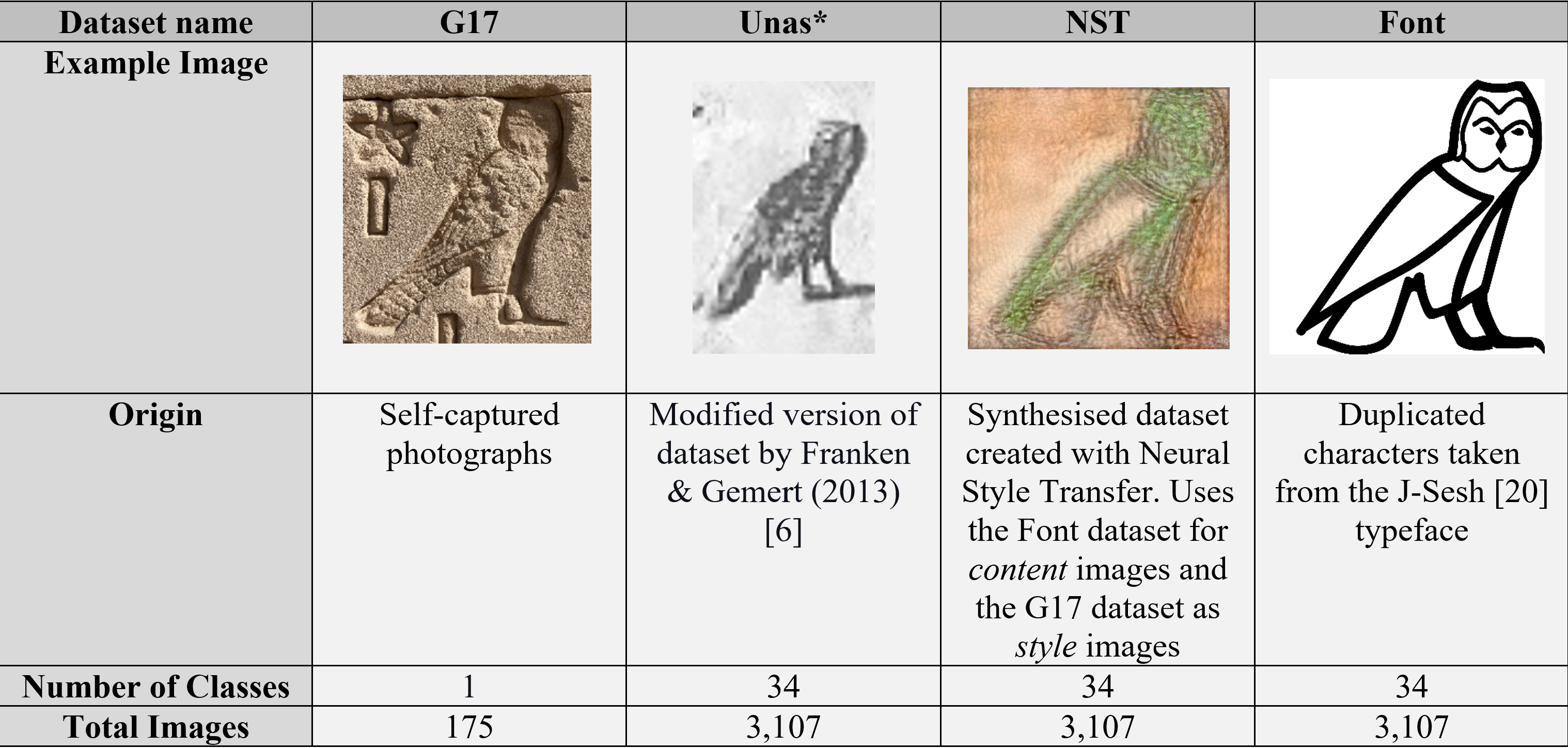}
    \caption{Table outlining key information from the various datasets used in the experiments.}
    \label{fig:dataset_breakdown}
\end{figure}

\section{Synthesising a Dataset with Neural Style Transfer}

\subsection{Neural Style Transfer Overview}

Neural Style Transfer (NST), as proposed by Gatys et al. \cite{gatys2015neural}, uses Convolutional Neural Networks (CNNs) to extract image features. Through downsampling, CNNs process pixel colour values in progressively smaller image segments \cite{gatys2015neural}. Each layer of a CNN examines these segments at different resolutions, summarising pixel values using convolutional filters (or kernels) to detect features like edges and shapes \cite{gatys2015neural}. For example, a convolution may detect a horizontal edge in a particular region, assigning it a value the model can use to understand the whole image. After passing these values through neural network layers, the model can attempt to classify the image based on these feature values. Thus, the ML process seeks to optimise this guesswork by fine-tuning functions in each layer for greater classification prediction accuracy.

For NST, two images, a \textit{content} and a \textit{style} image, are passed to a CNN, which extracts the features from both images but does not attempt any classification \cite{gatys2015neural}. Instead, the NST process tries to match the features of the \textit{content} image with the  \textit{style} image using a Gram matrix \cite{gatys2015neural}. This process augments each layer of the \textit{content} image with the respective layer in the \textit{style} image, resulting in a transference of textures, patterns, and colours from the \textit{style} reference image \cite{gatys2015neural}. Because CNNs downsample, high-level features like object outlines remain largely intact during the NST process, ensuring that the distinct outlines of hieroglyphic characters persist inside image outputs. Since NST augmentation retains the hieroglyphic character's identifiable features (i.e. its outline), output images should be classifiable and usable as training data. Therefore, \textit{content} and \textit{style} images are required to synthesise an NST dataset of ancient Egyptian hieroglyphs.

\subsection{\textit{Style} images: The G17 Dataset}

\begin{figure}[H]
    \centering
    \includegraphics[width=\textwidth]{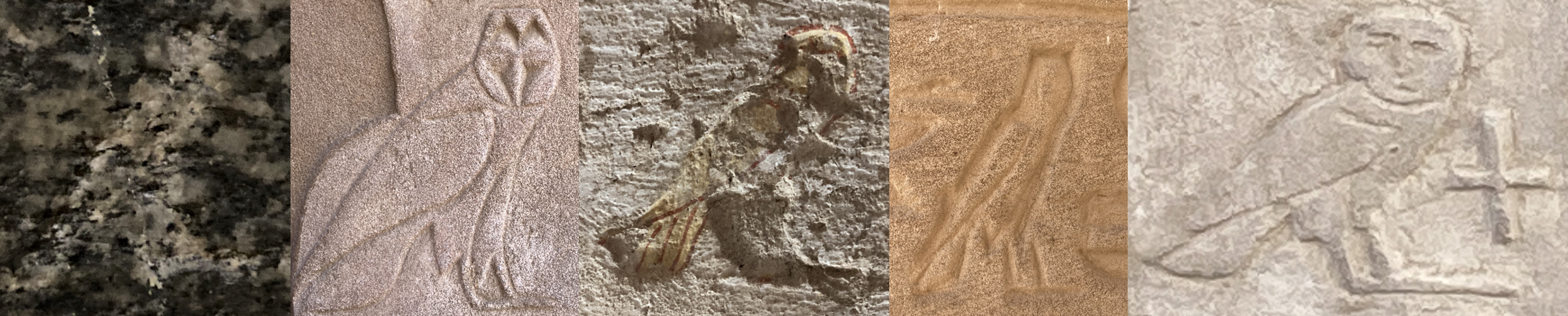}
    \caption{Five randomly selected examples of the G17 “owl” hieroglyph from the G17 dataset.}
    \label{fig:variety_styles}
\end{figure}

Theoretically, any image could be used as a style image to generate an NST dataset. Indeed, research suggests that picking various images can improve the heterogeneity of a synthesised dataset \cite{darma2020balinese}. However, I thought using photographs of real hieroglyphs would be befitting, as ancient Egyptian writing was written in many different artistic styles (see Figure \ref{fig:variety_styles}).

Conveniently, I could use self-collected images, which avoided any potential copyright issues. In December 2023, I studied in Egypt with the University of Auckland's Ancient History department. While there, I photographed the G17 "owl" hieroglyph at every archaeological site we visited to capture a snapshot of the diverse art styles used across the ancient Egyptian period. In total, I captured 175 different owls. The images were collected using an iPhone SE 2020. Some post-processing was then applied by converting the file format to turn the images into a usable dataset. Specifically, the file types were converted from .HEIC to .PNG and they were cropped to focus solely on the G17 hieroglyph. This collection of photographs will be referred to as the G17 dataset. The full dataset is available to view and download \href{https://drive.google.com/file/d/1PM847M2pcvaPBNeejqtHajSc03doyS8B/view?usp=sharing}{here}.

It is important to note that the range of art styles represented in the G17 dataset are not complete. There is also no comprehensive categorisation framework for the ancient Egyptian language's various art styles or fonts. Thus, there is no easy way to analyse or quantify the diversity of the examples in the G17 dataset apart from the human eye.

\subsection{\textit{Content} images: The J-Sesh Typeface}

Several digital typefaces for the ancient Egyptian language are available online. The typeface from the J-Sesh hieroglyphic editor \cite{jsesh2024jsesh} was chosen for generating the experimental NST dataset because it is open-source and popular with Egyptologists \cite{jseshwebsite}. The J-Sesh typeface is vector-based and uses .SVG files, so the individual characters were converted to a raster-based image format (.PNG) for CNN machine learning classification, as the models cannot process vector-based images.

A dataset of images was created using the J-Sesh typeface, hereafter referred to as the Font dataset. It was created as a baseline to assess the effects of NST augmentation. The Font dataset is highly homogenous, with each class containing only one example image duplicated to match the number of examples inside the Unas* and NST datasets. The Font dataset is available to view and download \href{https://drive.google.com/file/d/1PM847M2pcvaPBNeejqtHajSc03doyS8B/view?usp=sharing}{here}.

\subsection{Generating the Neural Style Transfer (NST) dataset}

\begin{figure}[ht]
    \centering
    \includegraphics[width=0.9\textwidth]{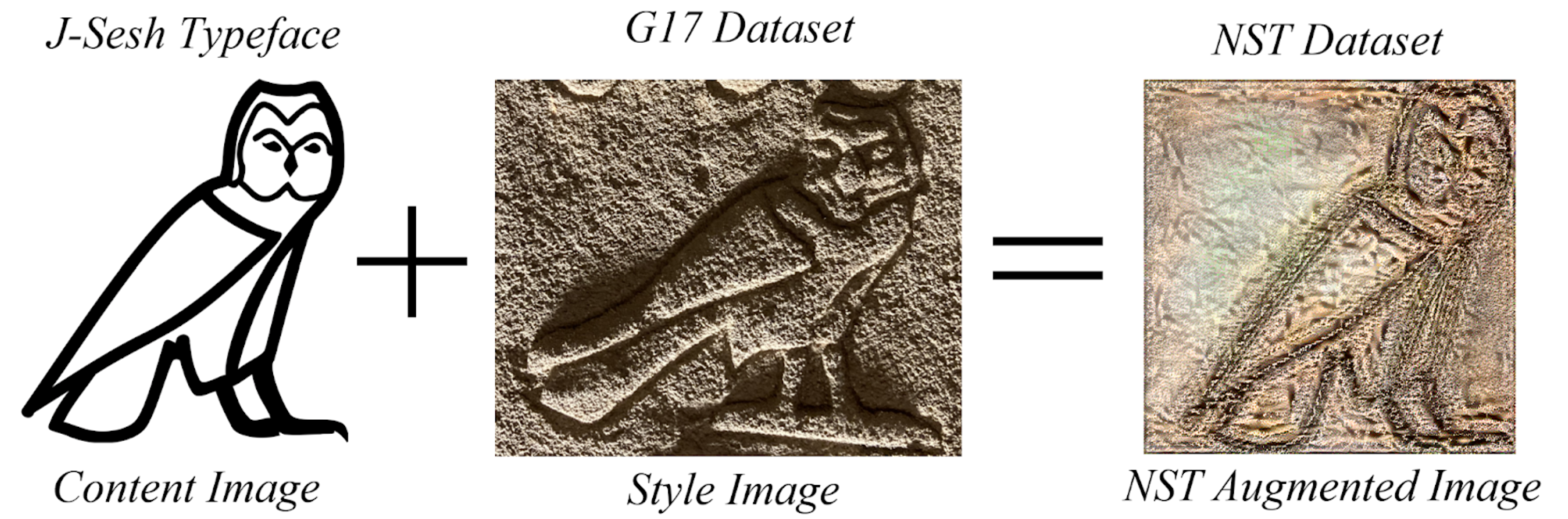}
    \caption{Example of \textit{content} and \textit{style} input images with the resultant NST-augmented output image.}
    \label{fig:nst_process}
\end{figure}

To generate the Neural Style Transfer (NST) dataset, Google’s TensorFlow \cite{tensorflow2024styletransfer} off-the-shelf Python-coded Jupyter Notebook implementation was chosen. The default hyper-parameters were used. Their code was refactored to allow for batch generation of NST images by exchanging the \textit{content} images but keeping the same \textit{style} image until an entire batch was completed. Then, new weights would be generated for a new \textit{style} image, and the process would be repeated (see Figure \ref{fig:nst_process}). 

As outlined earlier, the NST process attempts to transfer the texture and colour of an input \textit{style} image onto the form of an input \textit{content} image \cite{gatys2015neural}. To create the NST-generated dataset of ancient Egyptian hieroglyphics, the G17 dataset was used for the \textit{style} image inputs, and the J-Sesh typeface \cite{jsesh2024jsesh} was used for the \textit{content} image inputs. The G17 dataset was chosen to provide the \textit{style} inputs for the NST process because it contains various textures and colours depicting authentic ancient Egyptian hieroglyphs.

\begin{figure}[ht]
    \centering
    \includegraphics[width=0.9\textwidth]{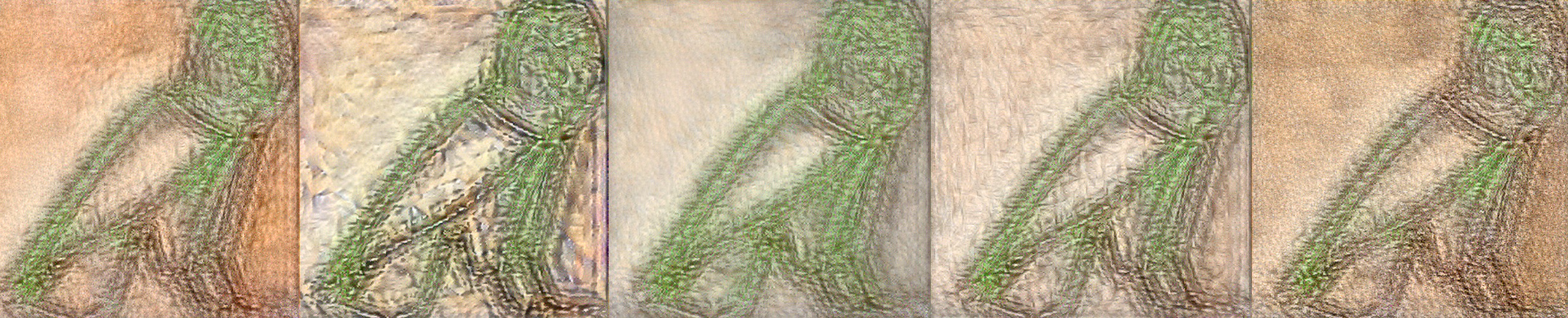}
    \caption{Examples of NST synthesised images for the G17 class of hieroglyphs. Note the underlying form from the same \textit{content} image persists despite being augmented by differing \textit{style} images.}
    \label{fig:nst_output}
\end{figure}

In total, 175 NST-augmented variations of 34 classes of hieroglyphs were generated (see Figure \ref{fig:nst_output} for some example outputs). Each image inside each class was made with one \textit{content} image taken from the J-Sesh \cite{jsesh2024jsesh} typeface and a different pairing of \textit{style} images taken from the G17 dataset. Each image took approximately 90 seconds to generate with my computer's hardware. To ensure the overall distribution of examples across the classes in the NST, Unas* and Font datasets were equal for comparability, excess images were removed with a Python script using the in-built random sample function. The NST dataset is available to view and download \href{https://drive.google.com/file/d/1PM847M2pcvaPBNeejqtHajSc03doyS8B/view?usp=sharing}{here}.

\section{Pruning the Unas Dataset}

The Unas dataset \cite{franken2013automatic} needed to be slightly amended before being used in my experiments. This altered version of the Unas dataset will be referred to as the Unas* dataset.

\begin{figure}[H]
    \centering
    \includegraphics[width=\textwidth]{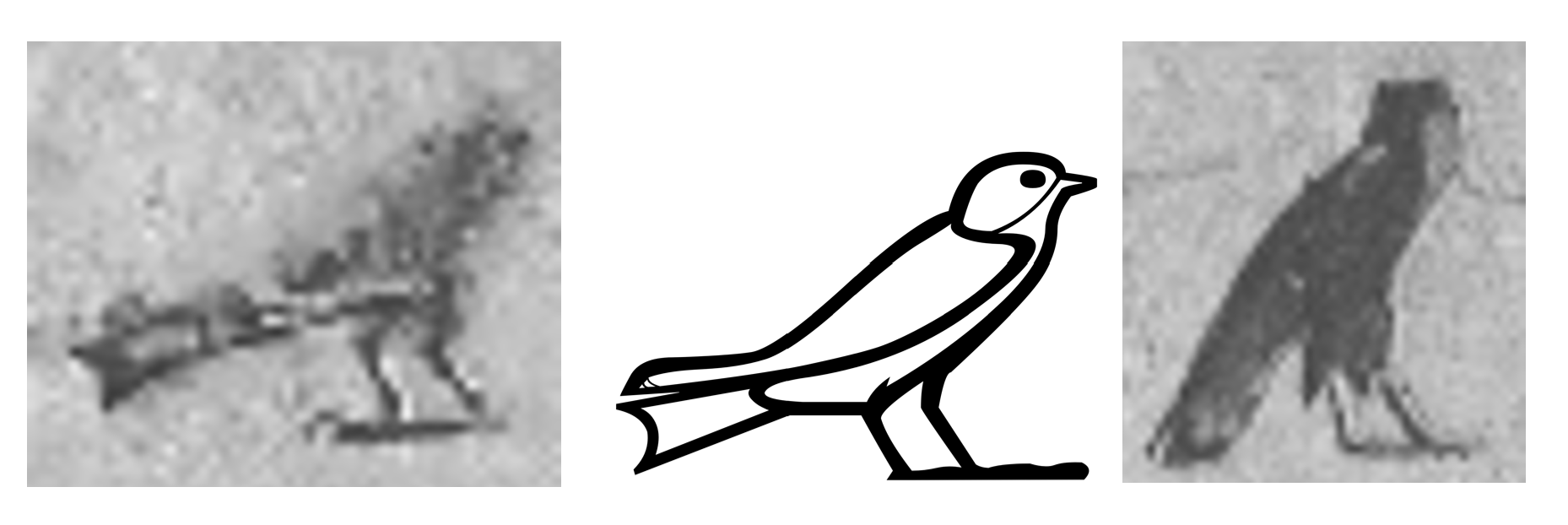}
    \caption{Left: The example ‘090361\_G17’ image from Unas dataset \cite{franken2013automatic} is mislabelled as a G17 “owl” when it is a G36 “swallow”. Centre: The G36 “swallow” hieroglyph as it appears in the J-Sesh typeface \cite{jsesh2024jsesh}. Right: A correctly classified G17 “owl” from the Unas dataset.}
    \label{fig:unas_birds}
\end{figure}

First, I reorganised the Unas dataset into a file hierarchy matching Gardiner’s sign list classification system, with 26 superclasses and numerous subclasses for the 171 individual types of hieroglyphic characters present in the Unas dataset \cite{franken2013automatic}. Next, because the authors of the Unas dataset acknowledged there could have been some mislabelled hieroglyphs, I manually checked their dataset, looking for any mislabelled hieroglyphs. My survey found numerous mislabelled examples which I corrected, such as the image titled ‘090361\_G17’, which was mislabelled as a G17 “owl” hieroglyph despite having the distinct crescent-shaped tail of a G36 “swallow” (see Figure \ref{fig:unas_birds}). It is unclear if other researchers have noticed these mistakes before as these errors have not been mentioned in other literature, which depends on the Unas dataset.

\section{Choosing the Image Classification Model: GlyphNet}

Although numerous image classifier models are available, I used the GlyphNet model presented by Barucci et al. \cite{barucci2021deep}. This was partly because GlyphNet has an accompanying paper published in a reputable journal (IEEE Access), and it is open source and available for download on GitHub \cite{barucci2021deep}. The authors also found that GlyphNet outperformed other models (ResNet-50, Inception-v3 and Xception) \cite{barucci2021deep}.

The GlyphNet architecture is a simplified and streamlined version of the Xception CNN \cite{barucci2021deep} architecture, which does not rely on pre-trained weights or transfer learning. As such, GlyphNet has 24 layers (12 less than Xception) and 498856 parameters inside the network (40 times less than Xception) \cite{barucci2021deep}. GlyphNet is written in Python code and runs inside a Jupyter Notebook.

\subsection{GlyphNet Hyper-parameters}

I used the hyper-parameters from Barucci et al. \cite{barucci2021deep} as a starting point for my approach.

Since multiple classes of hieroglyphs need to be classified by GlyphNet, categorical cross-entropy loss was used to adjust the functions across the various predicted classes \cite{barucci2021deep}. GlyphNet is built on the Keras ML library \cite{keras2024}; as such, the inbuilt Adam optimiser function was used to perform gradient descent. This algorithm efficiently improves the network’s classification functions based on the loss values \cite{barucci2021deep}. The learning rate was set to 0.001, based on Barucci et al. \cite{barucci2021deep}. However, since I was testing multiple datasets (Unas*, NST and Font), I decided to introduce early stopping after the first epoch when the loss value decreases by less than 0.05 to prevent overfitting.

However, introducing early stopping meant no training period lasted more than ten epochs. This diverged from Barucci et al.’s experimental settings, which used a scheduler to half the learning rate every 15 epochs over a fixed 100-epoch training period \cite{barucci2021deep}. Thus, I adjusted this hyper-parameter so that the learning rate was reduced by half every two epochs, proportionally matching the rate of decrease, which would occur across 100 epochs in a 10-epoch training period.

The resolutions of all images were also adjusted to 100x100 to be fed into the GlyphNet model as per Barucci et al.’s recommendation \cite{barucci2021deep}. The original resolution for the Unas* dataset was 75x50 and 512x512 for the NST and Font datasets. The images from the G17 dataset have various resolutions because they are cropped.

\section{Dataset Splits}

Three different splits of the NST and Unas* datasets were used for training to obtain averages and standard deviations for performance metrics. The GlyphNet models trained on these splits are hereafter referred to as NST (Split 1), NST (Split 2), NST (Split 3), Unas* (Split 1), Unas* (Split 2), and Unas* (Split 3), respectively. Following Barucci et al. \cite{barucci2021deep}, each split used a 75:15:15 ratio between the training, testing and validation subsets. These splits were randomly selected using a Python script with the built-in random sample function, which provided a different seed each time it was executed. However, only one model was trained on the Font dataset; since each class inside the dataset contains identical examples, any splits would also be identical regardless of the random function or the seed.

\section{Affine Data Augmentation}

Barucci et al. \cite{barucci2021deep} applied the same affine data-augmentation methods, including random translations along the x and y axis, random rotations within 10 degrees, random zooming between 0.95 and 1.05, and random horizontal flipping \cite{barucci2021deep}. This doubled the size of every class in each dataset before training and testing, bringing the total quantity of images in each training dataset to 6214.

\section{Software \& Hardware Specifications}

All scripts and experiments were coded in Python and run in a Jupyter Notebook inside an Anaconda environment \cite{anaconda2024}. A Windows 10 operating system computer was used with an AMD Ryzen 5 3600 6-Core CPU, NVIDIA GeForce RTX 2060 Super GPU and 16 GB RAM. Due to the available memory, all experiments' batch size was set to 32. The combined training and testing time for each model was approximately 1 minute.

\chapter{Results}

\section{Training, Validation and Testing}

\begin{figure}[H]
    \centering
    \includegraphics[width=\textwidth]{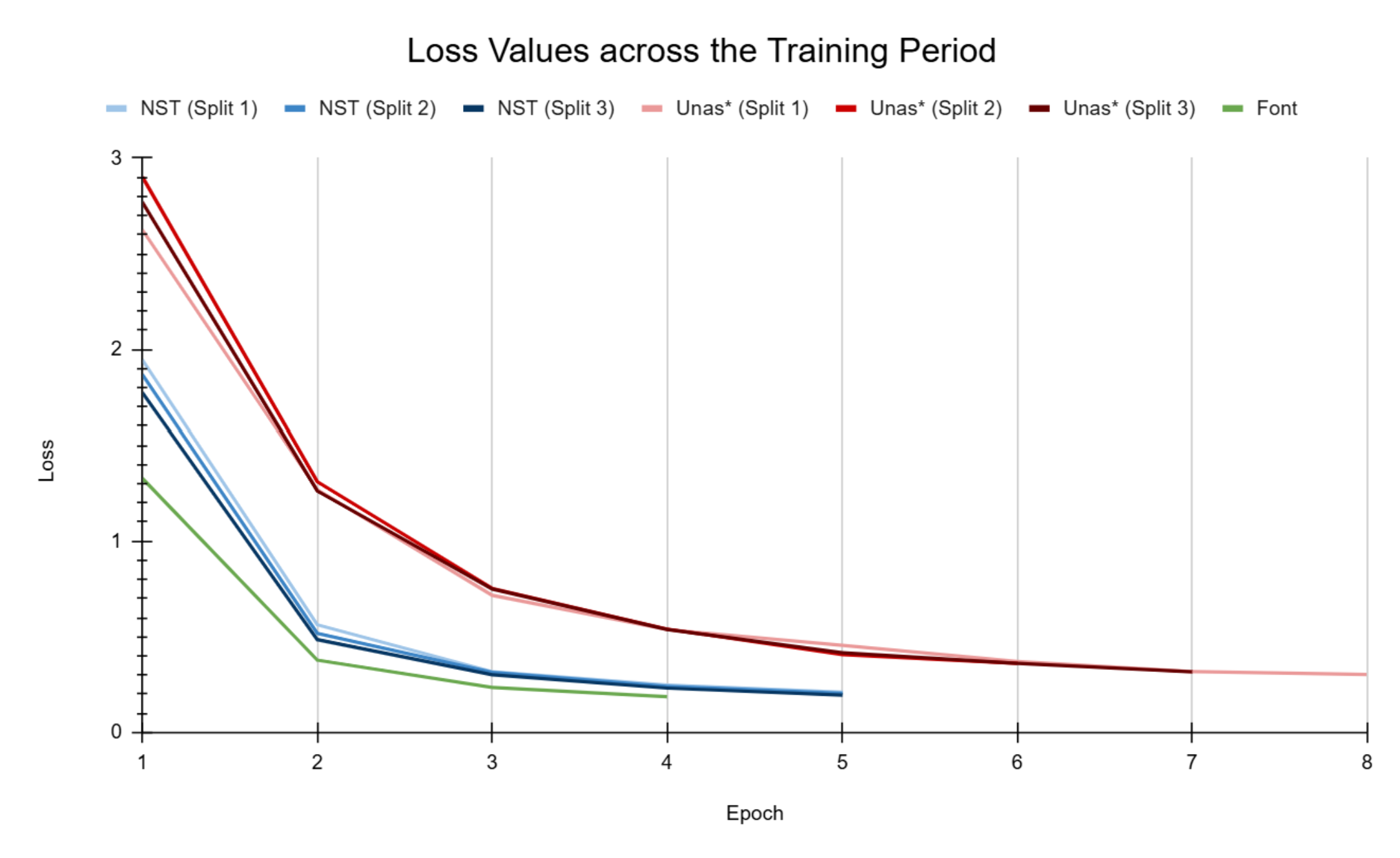}
    \caption{Loss function values are plotted over each epoch for each model trained. Note: Early stopping was used to reduce overfitting. Each dataset is grouped in similar colours to show how the GlyphNet \cite{barucci2021deep} model responded to the different data types.}
    \label{fig:loss}
\end{figure}

\begin{figure}[H]
    \centering
    \includegraphics[width=\textwidth]{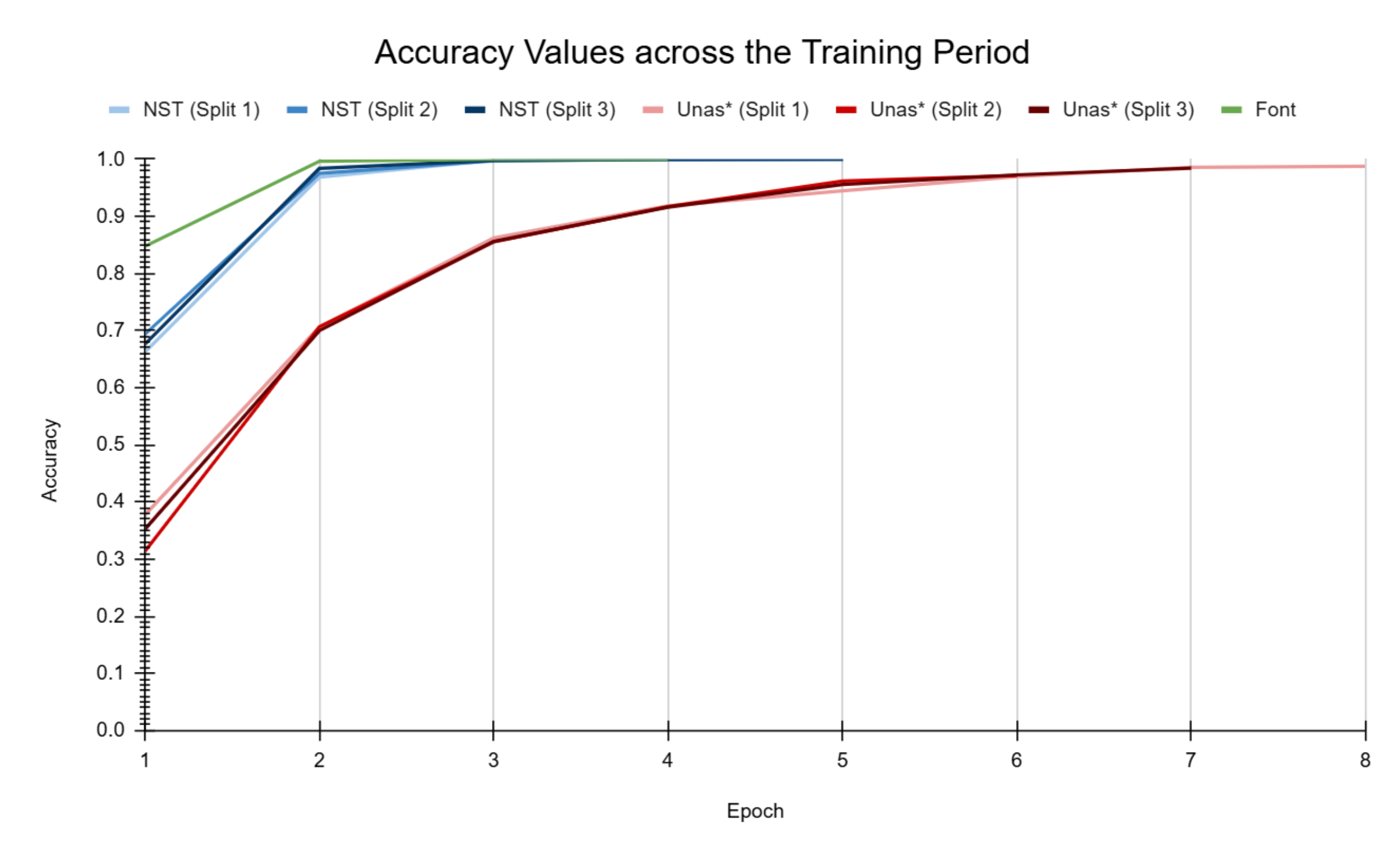}
    \caption{Training accuracy score during training plotted over each epoch for each model trained. Note: Early stopping was used to reduce overfitting. Each dataset is grouped in similar colours to show how the GlyphNet \cite{barucci2021deep} model responded to the different data types.}
    \label{fig:accuracy}
\end{figure}

Seven different versions of the GlyphNet model were created via training: three were derived from the Unas* dataset, three from the NST dataset, and one from the Font dataset. Class weighting was used to help reduce any issues arising from the unbalanced distribution of images across the classes (see Figure \ref{fig:class_distribution}); as such, macro average values are presented in this paper. I observed that the GlyphNet \cite{barucci2021deep} model responded differently depending on the type of image data it was working with. This is visualised by the loss and accuracy values in the graphs in Figures \ref{fig:loss} and \ref{fig:accuracy},  grouped into discrete curves depending on the dataset. Notably, the GlyphNet model consistently took twice as long to begin overfitting when trained on the images from the Unas* dataset than it would for the NST and Font datasets. This suggests that the Unas* dataset was the least homogenous considered, followed by the NST dataset and then the Font dataset.

The seven models were tested against unseen examples from their respective datasets (see Table \ref{tab:1} for results). The weights trained on the NST and Unas* datasets performed highly on their respective data types. This helps address the research questions, demonstrating that a dataset synthesised via NST from a digital typeface can be used to train a classification model for ancient Egyptian hieroglyphs.

\begin{table}[ht]
\centering
\caption{ Average results for the models tested on their respective test dataset splits.}
\renewcommand{\arraystretch}{1.5}  
\begin{tabular}[t]{p{1.72cm} p{1.72cm} p{2cm} p{2cm} p{2cm} p{2cm}}
\hline
Training Data & Testing Data & Accuracy & Precision & Recall & F1 Score \\
\hline
\textbf{NST} & \textbf{NST} & 0.99 \newline ($\pm$0.002) & 0.99 \newline ($\pm$0.001) & 0.99 \newline ($\pm$0.003) & 0.99 \newline ($\pm$0.002) \\
\textbf{Unas*} & \textbf{Unas*} & 0.94 \newline ($\pm$0.008) & 0.93 \newline ($\pm$0.01) & 0.99 \newline ($\pm$0.003) & 0.93 \newline ($\pm$0.01) \\
\textbf{Font} & \textbf{Font} & 0.058 \newline ($\pm$0) & 0.0039 \newline ($\pm$0) & 0.058 \newline ($\pm$0) & 0.0071 \newline ($\pm$0) \\
\hline
\end{tabular}
\label{tab:1}
\end{table}

Furthermore, GlyphNet models trained on the NST dataset outperformed the non-augmented Font dataset by 93.5\% on average. These results strongly indicate that NST, as a data augmentation tool, has a tangible positive effect on the quality and usefulness of the image data. In this context, NST has turned demonstrably unusable homogenous data into a usable dataset, performing on par with real photographic data (i.e. the Unas* dataset).

\section{Transferability}

To gauge transferability, models obtained from the Unas*, NST and Font datasets were first tested on unseen data, specifically the testing subsets from the other datasets (see Table \ref{tab:2} for results). None of the models performed particularly well at these tasks overall, although the models trained on the Unas* dataset consistently scored the highest across all the tasks. Like the first test in Table \ref{tab:1}, models trained on the NST dataset consistently outperformed the Font dataset model.

\begin{table}[ht]
\centering
\caption{Transferability results for the models against the other experimental datasets.}
\renewcommand{\arraystretch}{1.5}  
\begin{tabular}[t]{p{2cm} p{2cm} p{2cm} p{2cm} p{2cm} p{2cm}}
\hline
Training Data & Testing Data & Accuracy & Precision & Recall & F1 Score \\
\hline
\textbf{Unas*} & \textbf{NST} & 0.17 \newline ($\pm$0.04) & 0.16 \newline ($\pm$0.01) & 0.16 \newline ($\pm$0.02) & 0.19 \newline ($\pm$0.01) \\
\textbf{Font} & \textbf{NST} & 0.044 \newline ($\pm$0.005) & 0.002 \newline ($\pm$0.0003) & 0.033 \newline ($\pm$0.008) & 0.0039 \newline ($\pm$0.0006) \\
\textbf{NST} & \textbf{Unas*} & 0.059 \newline ($\pm$0.01) & 0.048 \newline ($\pm$0.03) & 0.05 \newline ($\pm$0.01) & 0.035 \newline ($\pm$0.01) \\
\textbf{Font} & \textbf{Unas*} & 0.045 \newline ($\pm$0.0001) & 0.0013 \newline ($\pm$0) & 0.029 \newline ($\pm$0) & 0.0025 \newline ($\pm$0) \\
\textbf{Unas*} & \textbf{Font} & 0.36 \newline ($\pm$0.08) & 0.24 \newline ($\pm$0.1) & 0.36 \newline ($\pm$0.06) & 0.27 \newline ($\pm$0.1) \\
\textbf{NST} & \textbf{Font} & 0.19 \newline ($\pm$0.2) & 0.061 \newline ($\pm$0.06) & 0.11 \newline ($\pm$0.1) & 0.072 \newline ($\pm$0.07) \\
\hline
\end{tabular}
\label{tab:2}
\end{table}

However, the images used in the transferability tests outlined in Table \ref{tab:2} are not representative of real ancient Egyptian hieroglyphs. The NST and Font datasets are completely digital and artificial, and the Unas* dataset is taken from grayscale photography. Real ancient Egyptian hieroglyphs were depicted in various colours, shapes, and sizes, made with different tools and materials.

Thus, a more realistic transferability test was conducted to better understand the capabilities of these weights at classifying real hieroglyphs. This was done using the G17 dataset, containing 175 photographs of diverse depictions of the “G17” owl hieroglyph from ancient Egyptian archaeological sites. Conveniently, G17 is one of the classes the models were trained to recognise and is well represented in all the datasets with a maximum class size of 175 examples. Since only one class is present in this dataset, the precision, recall and F1 score performance metrics are distorted; however, the accuracy metric is not. The results of this test are recorded in Table \ref{tab:3}.

The experimental results in Table \ref{tab:3} found that NST-augmented hieroglyphs can train a hieroglyphic classification model with high transferability to unseen images of real hieroglyphs. Once again, the models trained on the NST dataset massively outperformed the non-augmented Font dataset, which had no transferability.

\begin{table}[ht]
\centering
\caption{Transferability performance of the various GlyphNet models trained on the NST, Unas* and Font datasets on the G17 dataset.}
\renewcommand{\arraystretch}{1.5}  
\begin{tabular}[t]{p{2cm} p{2cm} p{2cm} p{2cm} p{2cm} p{2cm}}
\hline
Training Data & Testing Data & Accuracy & Precision & Recall & F1 Score \\
\hline
\textbf{NST} & \textbf{G17} & \textbf{0.74} \newline \textbf{($\pm$0.06)} & \textbf{0.092} \newline \textbf{($\pm$0.023)} & \textbf{0.069} \newline \textbf{($\pm$0.01)} & \textbf{0.079} \newline \textbf{($\pm$0.02)} \\
Unas* & G17 & 0.33 \newline ($\pm$0.2) & 0.13 \newline ($\pm$0.01) & 0.047 \newline ($\pm$0.04) & 0.064 \newline ($\pm$0.05) \\
Font & G17 & 0 \newline ($\pm$0) & 0 \newline ($\pm$0) & 0 \newline ($\pm$0) & 0 \newline ($\pm$0) \\
\hline
\end{tabular}
\label{tab:3}
\end{table}

\begin{table}[ht]
\centering
\caption{Transferability performance of the original GlyphNet model as published by Barucci et al. \cite{barucci2021deep} on the experimental datasets presented in this paper.}
\renewcommand{\arraystretch}{1.5}  
\begin{tabular}[t]{p{2cm} p{2cm} p{2cm} p{2cm} p{2cm} p{2cm}}
\hline
Training Data & Testing Data & Accuracy & Precision & Recall & F1 Score \\
\hline
\textbf{\cite{barucci2021deep}}& \textbf{G17} & \textbf{0.74} \newline \textbf{($\pm$0)} & \textbf{0.09} \newline \textbf{($\pm$0)} & \textbf{0.068} \newline \textbf{($\pm$0)} & \textbf{0.077} \newline \textbf{($\pm$0)} \\
\cite{barucci2021deep} & NST & 0.56 \newline ($\pm$0.002) & 0.49 \newline ($\pm$0.005) & 0.49 \newline ($\pm$0.01) & 0.45 \newline ($\pm$0.01) \\
\cite{barucci2021deep} & Unas* & 0.94 \newline ($\pm$0.01) & 0.90 \newline ($\pm$0.008) & 0.89 \newline ($\pm$0.02) & 0.89 \newline ($\pm$0.01) \\
\cite{barucci2021deep} & Font & 0.67 \newline ($\pm$0) & 0.64 \newline ($\pm$0) & 0.70 \newline ($\pm$0) & 0.66 \newline ($\pm$0) \\
\hline
\end{tabular}
\label{tab:4}
\end{table}

To independently verify the transferability scores in Tables \ref{tab:2} and \ref{tab:3}, the original state-of-the-art GlyphNet model produced by Barucci et al. \cite{barucci2021deep} was tested against the various experimental datasets (see Table \ref{tab:4} for results). Barucci et al.’s original model \cite{barucci2021deep} demonstrated superior transferability among the training datasets (compare Tables \ref{tab:3} and \ref{tab:4}). However, the models trained on the NST dataset obtained an accuracy score on par with the model published by Barucci et al. \cite{barucci2021deep}, which was trained on images of real hieroglyphs, primarily from the original Unas dataset (compare Tables \ref{tab:3} and \ref{tab:4}).

\chapter{Limitations}

\begin{figure}[H]
    \centering
    \includegraphics[width=0.8\textwidth]{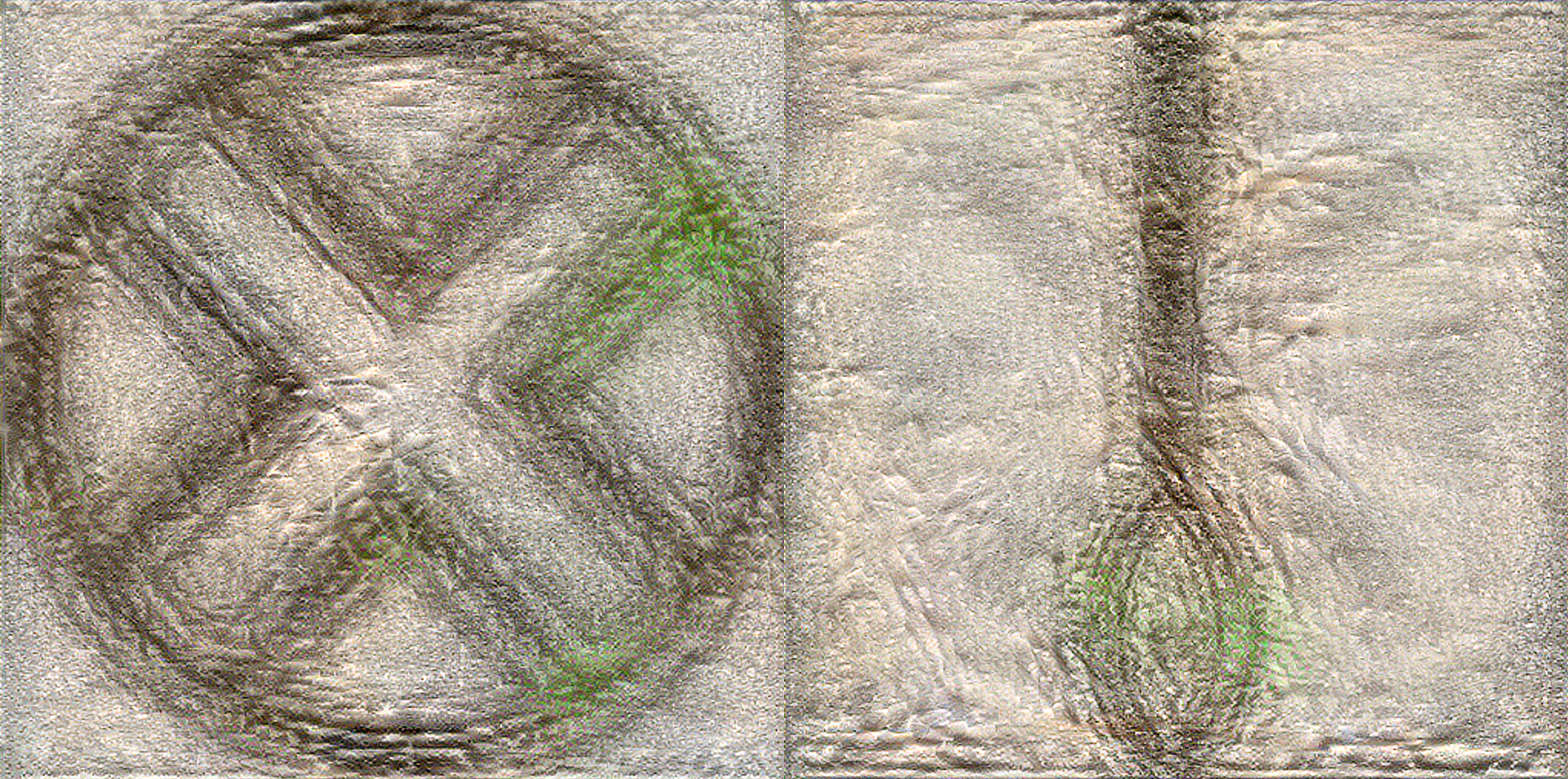}
    \caption{Left: Example O49 “city-plan” hieroglyph from the NST dataset, notice the distinct circle and ‘X’ shape. Right: Example P8 “oar” hieroglyph from the NST dataset. This P8 hieroglyph was generated after the O49 example on the left and the circle and ‘X’ shape can be faintly seen in the background.}
    \label{fig:burn_in}
\end{figure}

One limitation was that TensorFlow’s NST implementation \cite{tensorflow2024styletransfer} was not intended to produce batches of images as outlined in section 3.2. Although the code was able to be refactored to batch-produce NST images, doing so had the unintended side-effect of causing a slight watermark or burn-in effect, where outlines of previously processed hieroglyphs were visible in the background of subsequently generated hieroglyphs (see Figure \ref{fig:burn_in}.). The burn-in effect could be partially mitigated by providing a blank image as a \textit{content} image to the NST model before processing the rest of the images. Nevertheless, I could not completely resolve this issue, so future work should look into fixing this bug. It is unclear what effect this burn-in effect had on the performance of the NST models.

Another limitation of the scope of this research was hardware. Even though only a small subset (4.5\%) of the total number of ancient Egyptian hieroglyphs was needed for the experiments, it still took eight days to generate the NST dataset. Future work should seek to run the NST process on a more powerful computer to speed up dataset generation, permitting larger-sized experiments to be completed and faster experimental iterations.

\chapter{Discussion}

\section{Augmentation is the Key to Transferability}

Unsurprisingly, the results indicated that data augmentation methods significantly enhance image classification neural networks' transferability, which is already well-researched \cite[\dots]{atarsaikhan2017neural, mikolajczyk2018augmentation, mikolajczyk2019style,darma2020balinese, xiao2021progressive, li2023cross, mumuni2024survey}. For all of the models trained on the NST and Unas* datasets, it was observed that introducing affine data augmentation in the training process boosted transferability performance across all experiments. This boost improved the performance of the NST and Unas* models from random to intelligent guesses when tested against unseen data for transferability. This substantial increase deviates from the findings of Darma et al. \cite{darma2020balinese}, who found that affine augmentations only yielded a slight boost in performance overall. This affirms that affine augmentation is very important in the problem domain of classifying ancient Egyptian hieroglyphs.

Interestingly, affine augmentation did not result in a performance boost for the Font dataset, strongly suggesting that the additional layer of NST augmentation adds an essential degree of heterogeneity for such typeface data to become usable and more generalisable in ML classification tasks.

Furthermore, the consistently superior performance of NST augmented data over non-augmented data (i.e. the Font dataset) concurs with existing literature \cite{atarsaikhan2017neural ,mikolajczyk2018augmentation,mikolajczyk2019style,darma2020balinese,xiao2021progressive,li2023cross ,mumuni2024survey } that NST is a very useful data augmentation method for image processing and machine learning tasks.

\section{Hyper-parameters Make Comparison of Different Datasets Difficult}

The scores presented in Tables \ref{tab:1}, \ref{tab:2} and \ref{tab:3} are not definitive, nor should they be interpreted to suggest that NST augmented data is better than the Unas* dataset. Instead, the results in these tables address the research questions to explore if training a classification model on an NST synthesised dataset for ancient Egyptian hieroglyphs could be done and if such a model could match the transferability performance of a model trained on a photographic dataset of hieroglyphs (i.e. from the Unas pyramid texts \cite{franken2013automatic}).

\begin{table}[ht]
\centering
\caption{Summary of various experiments with different hyperparameters and settings. The accuracy result comes from testing the models with the G17 dataset transferability test. The settings that caused the GlyphNet model \cite{barucci2021deep} to perform better with the Unas* dataset than NST are bolded.}
\renewcommand{\arraystretch}{1.5}  
\begin{tabular}{lllr}
\hline
\textbf{Training} & \textbf{Learning} & \textbf{Early Stopping Condition } & \textbf{Accuracy} \\
\textbf{Dataset} & \textbf{Rate} &  & \\
\hline
NST & 0.001 &  Loss decreases by <0.1 & 0.22 ($\pm$0.19) \\
Unas & 0.001 &  Loss decreases by <0.1 & 0.10 ($\pm$0.06) \\
\textbf{NST} & \textbf{0.001} & \textbf{Loss decreases by <0.05} & \textbf{0.14} \textbf{($\pm$0.02)} \\
\textbf{Unas} & \textbf{0.001} & \textbf{Loss decreases by <0.05} & \textbf{0.32} \textbf{($\pm$0.29)} \\
NST & 0.0005 &  Loss decreases by <0.1 & 0.67 ($\pm$0.12) \\
Unas & 0.0005 &  Loss decreases by <0.1 & 0.08 ($\pm$0.09) \\
\textbf{NST} & \textbf{0.0005} & \textbf{Loss decreases by <0.05} & \textbf{0.24} \textbf{($\pm$0.03)} \\
\textbf{Unas} & \textbf{0.0005} & \textbf{Loss decreases by <0.05} & \textbf{0.56} \textbf{($\pm$0.40)} \\
NST & 0.0001 &  Loss decreases by <0.1 & 0.42 ($\pm$0.07) \\
Unas & 0.0001 &  Loss decreases by <0.1 & 0.01 ($\pm$0.01) \\
NST & 0.0001 &  Loss decreases by <0.05 & 0.26 ($\pm$0.02) \\
Unas & 0.0001 &  Loss decreases by <0.05 & 0.08 ($\pm$0.06) \\
\hline
\end{tabular}
\label{tab:5}
\end{table}

Systematic experimentation found that the GlyphNet model could exhibit superior performance with the Unas* or NST dataset, depending on the hyperparameters and settings (see Table \ref{tab:5}). In particular, I found that the early stopping condition had a large degree of sway on performance. This is because the model took longer to train on the Unas* dataset then the NST dataset. If the model finished earlier (when loss decreased by less than 0.1) the NST dataset had better performance as it was more optimally fitted. Whereas if the early stopping condition was later (i.e 0.05), the NST-trained models would become slightly overfitted, losing generalisability and performing worse than the Unas*-trained models, which were more optimally fitted.

Thus, it is difficult to compare the quality of the Unas/Unas* and NST datasets overall. The only conclusion I was able to draw between the two was that at a high level, the Unas dataset is the less homogenous dataset, as Figures \ref{fig:loss} and \ref{fig:accuracy} indicate the model overfitted slower than it did for the NST and Font datasets, indicating it took the model longer to learn, suggesting the data is less consistent/predictable (i.e. heterogenous). This is not that surprising given the photography in the Unas dataset consists of real hieroglyphs made by human hands, so a certain degree of variation is to be expected. However, since the Unas dataset was hand-annotated, expanding such a photographic dataset with more examples would be highly time-consuming \cite{franken2013automatic}. In contrast, while more homogenous (with the default settings), the NST synthesis method offers greater and faster possibilities, particularly around automatically generating complete datasets for the ancient Egyptian language.

While comparing the quality of datasets is interesting and worthy of inquiry, it is largely beside the point in a low-resource context. Instead, the best course of action is to merge existing and future datasets, which is exactly what other researchers have done previously \cite[\dots]{moustafa2022scriba, mohsen2023aegyptos}. Together, these datasets can synergise, increasing the overall heterogeneity of the available training data for classifying ancient Egyptian hieroglyphics.

\section{G17 Confusion Matrix Analysis}

\begin{figure}[H]
    \centering
\includegraphics[width=0.8\textwidth]{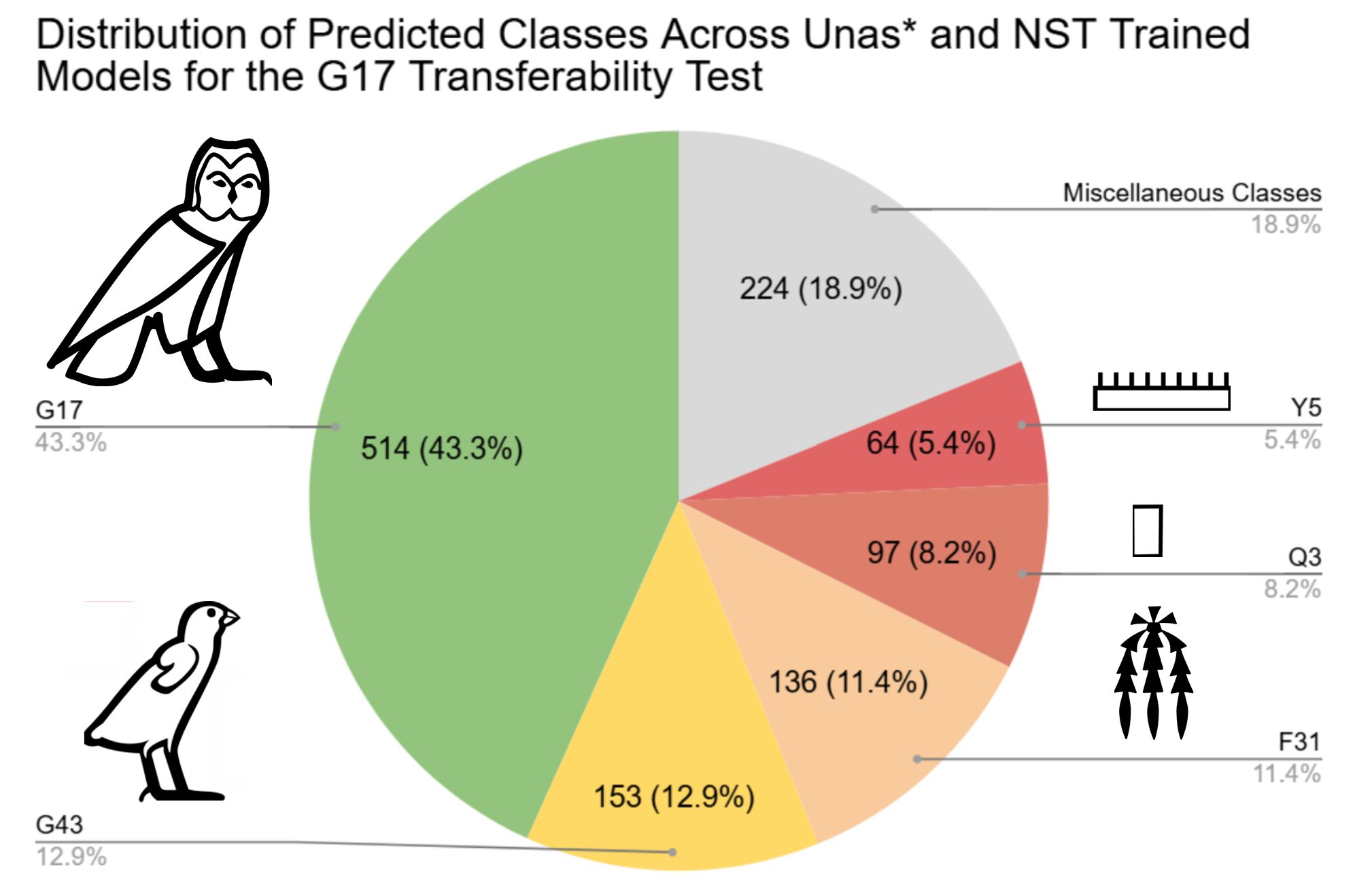}
    \caption{Pie chart depicting the distribution of predictions made by the models trained on the Unas* and NST datasets when classifying real photographs of ancient Egyptian hieroglyphs from the G17 dataset.}
    \label{fig:G_17}
\end{figure}

Although an in-depth class-by-class analysis of the various model's classification results is beyond the scope of this report, I thought it would be interesting to take a closer look at the results from the G17 transferability test since only one hieroglyph is present in that dataset. The results are presented in Figure \ref{fig:G_17}. 

Across the six Unas* and NST-trained models, they correctly identified the hieroglyph as G17 43.3\% of the time. The models most commonly mistook G17 for the G43 hieroglyph 12.9\% of the time. This is unsurprising as they belong to the same superclass (G) and both hieroglyphs depict birds.

\chapter{Future Work}

This paper demonstrates that digital ancient Egyptian hieroglyphic typefaces can be used successfully as datasets for training classification models by augmenting them with Neural Style Transfer. However, less than 5\% of the hieroglyphic characters that comprise the ancient Egyptian language were considered in these experiments. Therefore, the next logical step would be to use this novel NST method to generate a full dataset for the ancient Egyptian language. Currently, ther is no publicly available dataset containing all 765 classes of hieroglyphs in Gardiner's sign list \cite{gardiner1957hieroglyphic}. However, 765 is not an upper boundary for the number of classes a classification model could consider for ancient Egyptian characters: these are only the most common. For example, the J-Sesh typeface \cite{jsesh2024jsesh} contains 6,872 characters, including various variants.

Furthermore, the research here did not experiment with altering the various hyper-parameters for the NST process. These settings offer rich grounds for future experimentation. The hyper-parameters could likely be optimised or randomised to produce better, less homogeneous, results. I noted that while the texture of the NST images varied, the shape of the underlying \textit{content} image, the figure of the hieroglyph itself, was barely altered between images, meaning a degree of homogeneity remained in the dataset (see Figure \ref{fig:nst_output}). This homogeneity is also attested to in Figures \ref{fig:loss} and \ref{fig:accuracy}, where the loss and accuracy curves for the NST dataset are closer to matching the entirely homogeneous Font dataset than the Unas* dataset. Alongside altering the NST settings, future experiments could include applying affine augmentations to the \textit{content} images before the NST process to vary the underlying form of the hieroglyph, as was done in \cite{darma2020balinese}. Future work could also experiment with using images of different styles. An analysis must be made to assess if using photos of real Egyptian hieroglyphs, as was done with the G17 dataset, is the best practice. Perhaps images with unrelated styles and a greater variety of colours and textures could produce a more heterogeneous dataset.

\chapter{Conclusion}

In conclusion, I investigated the possibility of using Neural Style Transfer as a data augmentation method for generating image datasets from a typeface for machine learning classification. The results indicate such an NST-generated dataset can be used as a substitute for real photographic data in the training process for an image classification model with adjusted hyperparameters.  By utilising NST augmentation, near-limitless volumes of quality and transferable training data can be rapidly generated to cover the entirety of the ancient Egyptian language. As such, this novel approach promises to reduce, or trivialise, ancient Egyptian’s low resource language barriers, greatly opening up the domain for future and more complex machine learning research.

\appendix
\chapter{Appendix}

\section{Dataset Glossary}

\begin{table}[H]
    \centering
    \renewcommand{\arraystretch}{1.5}
    \begin{tabular}{p{0.24\linewidth}p{0.72\linewidth}}
\textbf{The Font dataset} & refers to the dataset made from characters taken from the J-Sesh typeface, duplicated to match the class distribution of the Unas* dataset. The Font dataset can be viewed and downloaded \href{https://drive.google.com/file/d/1PM847M2pcvaPBNeejqtHajSc03doyS8B/view?usp=sharing}{here}. \\

\textbf{The G17 dataset} & refers to the dataset of 175 real photos of the ancient Egyptian hieroglyph G17 “owl”, which were taken at Egyptian archaeological sites in 2023. The G17 dataset can be viewed and downloaded \href{https://drive.google.com/file/d/1PM847M2pcvaPBNeejqtHajSc03doyS8B/view?usp=sharing}{here}. \\

\textbf{The NST dataset} & refers to the dataset made using the Neural Style Transfer \cite{tensorflow2024styletransfer} data-augmentation process, merging the Font dataset as the \textit{content} images and the G17 dataset as the \textit{style} images. The NST dataset can be viewed and downloaded \href{https://drive.google.com/file/d/1PM847M2pcvaPBNeejqtHajSc03doyS8B/view?usp=sharing}{here}. \\

\textbf{The Unas* dataset} & refers to the altered Unas Pyramid Text Dataset taken from \cite{franken2013automatic} as outlined in the report. The Unas* dataset can be viewed and downloaded \href{https://drive.google.com/file/d/1PM847M2pcvaPBNeejqtHajSc03doyS8B/view?usp=sharing}{here}. \\

\textbf{The Unas dataset} & refers to the unaltered Unas Pyramid Text Dataset published by Franken and Gemert (2013) \cite{franken2013automatic}.\\

    \end{tabular}
 \label{tab:6}
\end{table}


\backmatter


\begin{thebibliography}{10}
\providecommand{\url}[1]{#1}
\csname url@samestyle\endcsname
\providecommand{\newblock}{\relax}
\providecommand{\bibinfo}[2]{#2}
\providecommand{\BIBentrySTDinterwordspacing}{\spaceskip=0pt\relax}
\providecommand{\BIBentryALTinterwordstretchfactor}{4}
\providecommand{\BIBentryALTinterwordspacing}{\spaceskip=\fontdimen2\font plus
\BIBentryALTinterwordstretchfactor\fontdimen3\font minus
  \fontdimen4\font\relax}
\providecommand{\BIBforeignlanguage}[2]{{%
\expandafter\ifx\csname l@#1\endcsname\relax
\typeout{** WARNING: IEEEtran.bst: No hyphenation pattern has been}%
\typeout{** loaded for the language `#1'. Using the pattern for}%
\typeout{** the default language instead.}%
\else
\language=\csname l@#1\endcsname
\fi
#2}}
\providecommand{\BIBdecl}{\relax}
\BIBdecl

\bibitem{template}
\BIBentryALTinterwordspacing
{Derek Lim} and {Peter Tan}. (2016) Honours thesis template. Accessed: Nov. 9,
  2024. [Online]. Available:
  \url{https://www.overleaf.com/articles/honours-thesis-template/ptcfpgjdhpkx}
\BIBentrySTDinterwordspacing

\bibitem{britannica2024}
\BIBentryALTinterwordspacing
{Editors of Encyclopaedia Britannica}. (2024, Oct) hieroglyph. Accessed: Nov.
  9, 2024. [Online]. Available:
  \url{https://www.britannica.com/topic/hieroglyph}
\BIBentrySTDinterwordspacing

\bibitem{britishmuseum2024rosetta}
\BIBentryALTinterwordspacing
{British Museum}. What is the rosetta stone? Accessed: Nov. 9, 2024. [Online].
  Available:
  \url{https://www.britishmuseum.org/blog/everything-you-ever-wanted-know-about-rosetta-stone}
\BIBentrySTDinterwordspacing

\bibitem{xinhua2024egypt}
\BIBentryALTinterwordspacing
{Xinhua News}. Egypt receives record-breaking 14.9 mln tourists in 2023.
  Accessed: Nov. 9, 2024. [Online]. Available:
  \url{https://english.news.cn/20240122/3df6ffe722474f1395177438709539c3/c.html}
\BIBentrySTDinterwordspacing

\bibitem{gardiner1957hieroglyphic}
A.~Gardiner, \emph{Egyptian Grammar}, 3rd~ed.\hskip 1em plus 0.5em minus
  0.4em\relax Cambridge, UK: Griffith Institute Oxford, 1957, ch. List of
  Hieroglyphic Signs, pp. 442--543.

\bibitem{franken2013automatic}
M.~Franken and J.~C.~V. Gemert, ``Automatic egyptian hieroglyph recognition by
  retrieving images as texts,'' in \emph{MM '13: Proceedings of the 21st ACM
  international conference on Multimedia}, 2013, pp. 765--768.

\bibitem{aneesh2024exploring}
N.~A. Aneesh, A.~Somasundaram, A.~Ameen, G.~S. Garimella, and R.~Jayashree,
  ``Exploring hieroglyph recognition: A deep learning approach,'' in \emph{2024
  2nd International Conference on Computer, Communication and Control (IC4)},
  2024, pp. 1--5.

\bibitem{guestnz2024hamilton}
\BIBentryALTinterwordspacing
{Guest New Zealand}. Visiting hamilton gardens waikato. Accessed: Nov. 9, 2024.
  [Online]. Available:
  \url{https://guestnewzealand.com/visiting-hamilton-gardens/}
\BIBentrySTDinterwordspacing

\bibitem{google2024translate}
\BIBentryALTinterwordspacing
{Google}. Google translate. Accessed: Nov. 9, 2024. [Online]. Available:
  \url{https://translate.google.com/}
\BIBentrySTDinterwordspacing

\bibitem{google2024fabricius}
\BIBentryALTinterwordspacing
{Google Arts \& Culture}. Fabricius. Accessed: Nov. 9, 2024. [Online].
  Available:
  \url{https://artsandculture.google.com/experiment/fabricius/gwHX41Sm0N7-Dw?hl=en}
\BIBentrySTDinterwordspacing

\bibitem{barucci2021deep}
{A. Barucci}, {C. Cucci}, {M. Franci}, {M. Loschiavo}, and {F. Argenti}, ``A
  deep learning approach to ancient egyptian hieroglyphs classification,''
  \emph{IEEE Access}, vol.~9, pp. 123\,438--123\,447, Sept. 2021.

\bibitem{lion2024unsupervised}
P.~Lion, E.~Trunz, and R.~Klein, ``Unsupervised detection and localization of
  egyptian hieroglyphs,'' in \emph{Eurographics Workshop on Graphics and
  Cultural Heritage}, M.~Corsini, D.~Ferdani, A.~Kuijper, and H.~Kutlu,
  Eds.\hskip 1em plus 0.5em minus 0.4em\relax The Eurographics Association,
  2024.

\bibitem{lombardi2024localisation}
L.~Lombardi, F.~Mercaldo, and A.~Santone, ``Egyptian hieroglyphs localisation
  through object detection,'' in \emph{Proc. 19th Int. Conf. Softw. Technol.},
  vol.~1.\hskip 1em plus 0.5em minus 0.4em\relax ICSOFT, SciTePress, 2024, pp.
  434--441.

\bibitem{cucci2024hyperspectral}
C.~Cucci \emph{et~al.}, ``Hyperspectral imaging and convolutional neural
  networks for augmented documentation of ancient egyptian artefacts,''
  \emph{Heritage Science}, vol.~12, no.~75, 2024.

\bibitem{gatys2015neural}
{L. A. Gatys}, {A. S. Ecker}, and {M. Bethge}, ``A neural algorithm of artistic
  style,'' \emph{arXiv}, Sept. 2015.

\bibitem{mikolajczyk2018augmentation}
A.~Mikołajczyk and M.~Grochowski, ``Data augmentation for improving deep
  learning in image classification problem,'' in \emph{2018 International
  Interdisciplinary PhD Workshop (IIPhDW)}, Świnouście, Poland, 2018, pp.
  117--122.

\bibitem{mikolajczyk2019style}
{A. Mikołajczyk} and {M. Grochowski}, ``Style transfer-based image synthesis
  as an efficient regularization technique in deep learning,'' in \emph{2019
  24th International Conference on Methods and Models in Automation and
  Robotics (MMAR)}, Miedzyzdroje, Poland, 2019, pp. 42--47.

\bibitem{darma2020balinese}
{I. W. A. S. Darma}, {N. Suciati}, and {D. Siahaan}, ``Neural style transfer
  and geometric transformations for data augmentation on balinese carving
  recognition using mobilenet,'' \emph{International Journal of Intelligent
  Engineering and Systems}, vol.~13, no.~6, pp. 348--363, 2020.

\bibitem{xiao2021progressive}
{Q. Xiao}, {B. Liu}, {Z. Li}, {W. Ni}, {Z. Yang}, and {L. Li}, ``Progressive
  data augmentation method for remote sensing ship image classification based
  on imaging simulation system and neural style transfer,'' \emph{IEEE Journal
  of Selected Topics in Applied Earth Observations and Remote Sensing},
  vol.~14, pp. 9176--9186, 2021.

\bibitem{jsesh2024jsesh}
\BIBentryALTinterwordspacing
W.~C. Serge J.P.~Thomas, Serge~Rosmorduc. Jsesh sources. Accessed: Nov. 9,
  2024. [Online]. Available:
  \url{https://github.com/rosmord/jsesh/tree/master/jseshGlyphs/src/main/resources/jseshGlyphs}
\BIBentrySTDinterwordspacing

\bibitem{unicode2024hieroglyphs}
\BIBentryALTinterwordspacing
{Unicode}. Egyptian hieroglyphs. Accessed: Nov. 9, 2024. [Online]. Available:
  \url{https://unicode.org/charts/nameslist/n_13000.html}
\BIBentrySTDinterwordspacing

\bibitem{atarsaikhan2017neural}
G.~Atarsaikhan, B.~K. Iwana, A.~Narusawa, K.~Yanai, and S.~Uchida, ``Neural
  font style transfer,'' in \emph{2017 14th IAPR International Conference on
  Document Analysis and Recognition (ICDAR)}, 2017, pp. 51--56.

\bibitem{li2023cross}
C.~Li, Y.~Taniguchi, M.~Lu, S.~Konomi, and H.~Nagahara, ``Cross-language font
  style transfer,'' \emph{Applied Intelligence}, vol.~53, pp. 18\,666--18\,680,
  2023.

\bibitem{mumuni2024survey}
A.~Mumuni, F.~Mumuni, and N.~K. Gerrar, ``A survey of synthetic data
  augmentation methods in computer vision,'' \emph{arXiv}, Mar. 2024.

\bibitem{duque2017deciphering}
{J. Duque-Domingo}, {P. J. Herrera}, {E. Valero}, and {C. Cerrada},
  ``Deciphering egyptian hieroglyphs: Towards a new strategy for navigation in
  museums,'' \emph{Sensors}, vol.~17, no.~3, p. 589, Mar. 2017.

\bibitem{elnabawy2018image}
{R. Elnabawy}, {R. Elias}, and {M. Salem}, ``Image based hieroglyphic character
  recognition,'' in \emph{14th International Conference on Signal-Image
  Technology \& Internet-Based Systems (SITIS)}, 2018, pp. 32--39.

\bibitem{moustafa2022scriba}
R.~Moustafa \emph{et~al.}, ``Hieroglyphs language translator using deep
  learning techniques (scriba),'' in \emph{2nd International Mobile,
  Intelligent, and Ubiquitous Computing Conference (MIUCC)}, 2022, pp.
  125--132.

\bibitem{mohsen2023aegyptos}
S.~E. Mohsen, R.~Mansour, A.~Bassem, B.~Dessouky, S.~Refaat, and T.~M. Ghanim,
  ``Aegyptos: Mobile application for hieroglyphs detection, translation and
  pronunciation,'' in \emph{2023 International Mobile, Intelligent, and
  Ubiquitous Computing Conference (MIUCC)}, 2023, pp. 1--8.

\bibitem{sobhy2023translator}
{A. Sobhy}, {M. Helmy}, {M. Khalil}, {S. Elmasry}, {Y. Boules}, and {N.
  Negied}, ``An ai based automatic translator for ancient hieroglyphic language
  — from scanned images to english text,'' \emph{IEEE Access}, vol.~11, pp.
  38\,796--38\,804, Apr. 2023.

\bibitem{guidi2023segmentation}
T.~Guidi \emph{et~al.}, ``Egyptian hieroglyphs segmentation with convolutional
  neural networks,'' \emph{Algorithms}, vol.~16, no.~2, p.~79, Feb. 2023.

\bibitem{custer2024}
\BIBentryALTinterwordspacing
M.~Custer. (2023) Cota\_coco\_anks computer vision project. Accessed: Nov. 9,
  2024. [Online]. Available:
  \url{https://universe.roboflow.com/matthew-custer-bclqa/cota_coco_anks}
\BIBentrySTDinterwordspacing

\bibitem{hieratic}
\BIBentryALTinterwordspacing
{Editors of Encyclopaedia Britannica}. (2024) Hieratic script. Accessed: Nov.
  9, 2024. [Online]. Available:
  \url{https://www.britannica.com/topic/hieratic-script}
\BIBentrySTDinterwordspacing

\bibitem{jseshwebsite}
S.~Rosmorduc. (2014) Jsesh documentation. https://jsesh.qenherkhopeshef.org/.
  Accessed: Nov. 9, 2024.

\bibitem{tensorflow2024styletransfer}
\BIBentryALTinterwordspacing
TensorFlow. Neural style transfer. Accessed: Nov. 9, 2024. [Online]. Available:
  \url{https://www.tensorflow.org/tutorials/generative/style_transfer}
\BIBentrySTDinterwordspacing

\bibitem{keras2024}
\BIBentryALTinterwordspacing
{Keras}. (2024) Adam. Accessed: Nov. 9, 2024. [Online]. Available:
  \url{https://keras.io/api/optimizers/adam/}
\BIBentrySTDinterwordspacing

\bibitem{anaconda2024}
\BIBentryALTinterwordspacing
{Anaconda}. (2024) Anaconda: The world's most popular data science platform.
  Accessed: Nov. 9, 2024. [Online]. Available: \url{https://www.anaconda.com/}
\BIBentrySTDinterwordspacing

\end{thebibliography}
\end{document}